\documentclass[11pt,a4paper]{article}
\usepackage[utf8]{inputenc}
\usepackage[T1]{fontenc}
\usepackage{lmodern}
\usepackage[margin=2.5cm]{geometry}
\usepackage{amsmath,amssymb,array,booktabs,longtable,graphicx,xcolor}
\usepackage[round]{natbib}
\usepackage[colorlinks=true,linkcolor=blue!50!black,citecolor=blue!50!black,urlcolor=blue!50!black]{hyperref}
\providecommand{\tightlist}{\setlength{\itemsep}{0pt}\setlength{\parskip}{0pt}}
\newcommand{\pkg}[1]{\textbf{#1}}
\newcommand{\CRANpkg}[1]{\href{https://CRAN.R-project.org/package=#1}{\pkg{#1}}}

\title{Fast and Memory-Light Evaluation of Density-Based Clusterings with the DISCO Index in R}
\author{Amin Entezari\\
\small Universit\`a di Milano-Bicocca, Milan, Italy\\
\small ORCID: \href{https://orcid.org/0009-0005-4733-6032}{0009-0005-4733-6032}\\
\small \href{mailto:amin_entezari@outlook.com}{\nolinkurl{amin_entezari@outlook.com}}}
\date{November 2026\\[0.5ex]\small Preprint.}

\begin{document}
\maketitle

\begin{abstract}
DISCO is a recent internal cluster validity index for density-based clustering that, unlike earlier indices, also evaluates the quality of noise labels. The discoCVI package makes DISCO available in R. This article describes the design of the package and a typical workflow, in which DISCO is used to compare clusterings and to choose the parameters of HDBSCAN without ground truth, followed by the implementation introduced in version 0.2.0. DISCO is defined through density-connectivity distances between all pairs of points, so direct implementations store several n-by-n matrices. We show that every quantity DISCO needs can be read off the single-linkage merge tree of the mutual-reachability minimum spanning tree, so that the index can be computed exactly with memory that grows linearly with the number of points, while the running time remains quadratic as in the original definition. The implementation is tested against the previous version, agrees with the original Python implementation to within \(3 \times 10^{-13}\), and corrects the score of singleton clusters to match the published definition. In benchmarks, discoCVI is 23 to 138 times faster than its previous version for 500 to 4,000 points and 3 to 9 times faster than the Python reference for 500 to 8,000 points, using a fraction of the memory of both, and it scores 100,000 points in about 50 seconds with less than 200 MB of memory.
\end{abstract}

\section{Introduction}\label{introduction}

Clustering without ground truth needs internal cluster validity indices
(CVIs), which score a partition using only the data and the labels. They are
used to compare the output of different algorithms and, above all, to choose
the parameters of a clustering algorithm when no labelled data exist. For
density-based algorithms such as DBSCAN \citep{ester1996dbscan} and HDBSCAN
\citep{campello2013hdbscan}, which find clusters of arbitrary shape and label
low-density points as noise, classical indices such as the silhouette
coefficient \citep{rousseeuw1987silhouette} are a poor fit: they favour convex
clusters and have no notion of noise. Density-based indices such as DBCV
\citep{moulavi2014dbcv} address the first problem, and the recent DISCO index
\citep{beer2026disco} also addresses the second: it scores every point labelled
as noise according to whether it really lies in a sparse region far from all
clusters.

\textbf{Existing software.} Several CRAN packages provide internal validity
indices: \CRANpkg{cluster} computes the silhouette coefficient,
\CRANpkg{clusterCrit} and \CRANpkg{fpc} offer a wide range of classical
indices (\CRANpkg{fpc} also implements the density-based index CDbw), and
\CRANpkg{DBCVindex} implements DBCV \citep{chicco2025dbcv}. None of them
evaluates noise labels. The original DISCO implementation is written in
Python and is not distributed as a package. \CRANpkg{discoCVI} \citep{discocvi}
makes DISCO available in R; its first release translated the reference
implementation line by line.

\textbf{The gap.} DISCO is defined through the density-connectivity distance
(dc-distance) \citep{beer2023dc} between every pair of points. Both the reference
implementation and \CRANpkg{discoCVI} 0.1.1 therefore build the full matrices
of mutual-reachability distances and dc-distances. Time and memory grow
quadratically with the number of points \(n\): version 0.1.1 needs about 19
seconds and 1.2 GB to score 4,000 points, and a few thousand points more
exhaust a laptop. Because an index is usually evaluated for many candidate
clusterings, this cost is multiplied by the size of the search, which matters
for running time and also for the energy use of the analysis
\citep{strubell2019energy, schwartz2020green, lannelongue2021green}.

\textbf{This article.} We describe the design of \CRANpkg{discoCVI} and a typical
workflow, and then the implementation introduced in version 0.2.0, which
computes DISCO exactly without any \(n \times n\) matrix. We show that this is
possible because the dc-distance between two points equals the height at
which they are merged in the single-linkage tree of the mutual-reachability
graph, so that all per-cluster sums and minima of dc-distances can be
accumulated in one pass over that tree. We report tests against the previous
version and against the original Python implementation, and a benchmark of
time, memory and energy.

\section{The DISCO index}\label{the-disco-index}

We follow the notation of \citet{beer2026disco}. Let
\(X = \{x_1, \dots, x_n\} \subset \mathbb{R}^p\) and let a clustering
\(\mathcal{C} = \{C_1, \dots, C_K\}\) consist of disjoint clusters, with the
remaining points \(N = X \setminus \bigcup_i C_i\) labelled as noise; for a
cluster point \(x\), \(\hat{C}_x\) denotes its cluster.

\textbf{Distances.} For a parameter \(\mu\) (\texttt{min\_points} in the package, default
5), the core distance \(\kappa(x)\) is the Euclidean distance from \(x\) to its
\(\mu\)-th nearest neighbour, where \(x\) counts as its own first neighbour, as
in the reference implementation; equivalently, it is the distance to the
\((\mu - 1)\)-th nearest \emph{other} point, and \(\mu \geq 2\) is required. The
mutual-reachability distance \citep{campello2013hdbscan} is
\[
d_m(x, y) = \max\{\kappa(x), \kappa(y), \lVert x - y \rVert\},
\]
and the dc-distance \citep{beer2023dc} is the minimax path distance in the
complete graph weighted by \(d_m\), that is, the smallest possible value of the
largest edge on a path from \(x\) to \(y\), with \(d_{dc}(x, x) = 0\). It equals the
largest edge on the path between \(x\) and \(y\) in any minimum spanning tree
(MST) of that graph \citep{hu1961maxcap}; in particular it does not depend on how
ties between equal edge weights are broken when the MST is built.

\textbf{DISCO score.} DISCO is defined pointwise, and the score of a clustering is
the mean of the pointwise scores,
\[
\rho(X) = \frac{1}{n} \sum_{x \in X} \rho(x), \qquad
\rho(x) = \begin{cases} \rho_{\text{cluster}}(x) & \text{if } x \in C_i
\text{ for some } i,\\ \rho_{\text{noise}}(x) & \text{if } x \in N. \end{cases}
\]

\textbf{Cluster points.} With \(\bar{d}_{dc}(x, C)\) the mean dc-distance from \(x\)
to the points of \(C\), a cluster point receives
\[
\rho_{\text{cluster}}(x) = \min_{C_i \neq \hat{C}_x}
\frac{\bar{d}_{dc}(x, C_i) - \bar{d}_{dc}(x, \hat{C}_x)}
     {\max\{\bar{d}_{dc}(x, C_i), \bar{d}_{dc}(x, \hat{C}_x)\}}.
\]
For fixed \(a \geq 0\) the function \(b \mapsto (b - a)/\max\{a, b\}\) is
non-decreasing, so the minimum is attained at the closest cluster, and
\(\rho_{\text{cluster}}\) is the silhouette coefficient computed on
dc-distances:
\begin{gather*}
\rho_{\text{cluster}}(x) = \frac{b(x) - a(x)}{\max\{a(x), b(x)\}}, \\
a(x) = \frac{1}{|\hat{C}_x| - 1} \sum_{y \in \hat{C}_x \setminus \{x\}} d_{dc}(x, y), \qquad
b(x) = \min_{C_i \neq \hat{C}_x} \frac{1}{|C_i|} \sum_{y \in C_i} d_{dc}(x, y).
\end{gather*}
As in the reference implementation, which uses the silhouette function of
scikit-learn \citep{pedregosa2011sklearn}, the within-cluster mean \(a(x)\) is taken
over the \emph{other} members of \(\hat{C}_x\).

\textbf{Noise points.} With the core-distance threshold of a cluster
\(\kappa(C) = \max_{y \in C} \kappa(y)\) and \(m_C(x) = \min_{y \in C} d_{dc}(x, y)\),
a noise point receives the minimum of two scores,
\[
\rho_{\text{sparse}}(x) = \min_{C \in \mathcal{C}} \frac{\kappa(x) - \kappa(C)}{\max\{\kappa(x), \kappa(C)\}}, \qquad
\rho_{\text{far}}(x) = \min_{C \in \mathcal{C}} \frac{m_C(x) - \kappa(C)}{\max\{m_C(x), \kappa(C)\}},
\]
\[
\rho_{\text{noise}}(x) = \min\{\rho_{\text{sparse}}(x), \rho_{\text{far}}(x)\}.
\]
\(\rho_{\text{sparse}}\) is positive if \(x\) is sparser than every cluster, and
\(\rho_{\text{far}}\) is positive if \(x\) is not density-connected to any
cluster.

\textbf{Edge cases} \citep[Section 3.1]{beer2026disco}. (i) If all points are noise,
every score is \(-1\). (ii) If there is a single cluster and no noise, every
score is \(0\). (iii) Points in singleton clusters score \(0\). (iv) If there is
a single cluster plus noise, the cluster points are compared with the closest
noise point instead of the (non-existent) closest cluster, i.e.~\(b(x)\) is
replaced by \(\min_{y \in N} d_{dc}(x, y)\). (v) If a denominator is zero, for
example when more than \(\mu\) duplicate points have core distance \(0\), the
corresponding fraction is set to \(0\).

\section{Package design}\label{package-design}

\textbf{Intended users.} \CRANpkg{discoCVI} is written for data analysts who
cluster data with a density-based method, for example with
\CRANpkg{dbscan} \citep{hahsler2019dbscan}, and need to judge or tune the result
without ground truth. Using the package requires no knowledge of how DISCO
is computed: one function returns a score, and a second returns one score
per point.

\textbf{Interface.} The package exports five functions with plain numeric inputs
and outputs:

\begin{itemize}
\tightlist
\item
  \texttt{disco\_score(X,\ labels,\ min\_points\ =\ 5)} returns the DISCO score of a
  clustering, a single number in \([-1, 1]\);
\item
  \texttt{disco\_samples(X,\ labels,\ min\_points\ =\ 5)} returns the pointwise scores
  \(\rho(x)\);
\item
  \texttt{p\_cluster()} and \texttt{p\_noise()} return the two components of the score for
  cluster points and noise points (\texttt{p\_noise()} returns \(\rho_{\text{sparse}}\)
  and \(\rho_{\text{far}}\) separately);
\item
  \texttt{compute\_dc\_distances(X,\ min\_points\ =\ 5)} returns the dc-distance matrix,
  which is useful for analysing small datasets or comparing many labellings
  of the same data.
\end{itemize}

\texttt{X} is a numeric matrix or data frame and \texttt{labels} an integer vector in
which \texttt{-1} marks noise. The function names, the argument \texttt{min\_points} and
the noise label \texttt{-1} are those of the reference Python implementation, which
itself follows the conventions of scikit-learn, so that code and results
carry over between the two languages.

\textbf{Design choices.} The functions return plain numeric vectors (or a list of
two vectors for \texttt{p\_noise()}) rather than objects of an S3, S4 or R6 class.
A validity index produces numbers that users typically put into a data
frame, average, plot, or pass to \texttt{sapply()} over a grid of parameters, and a
class would add no information while making these operations less direct.
The package is deliberately agnostic to the clustering algorithm: it accepts
labels from DBSCAN, HDBSCAN, OPTICS, \(k\)-means or any other source, and
therefore lists \CRANpkg{dbscan} only as a suggested package. The only
dependencies are \CRANpkg{FNN} \citep{fnn}, for the nearest-neighbour search that
defines the core distances, and \CRANpkg{Rcpp} \citep{eddelbuettel2011rcpp}, for
the compiled spanning-tree step. The computation is deterministic and
single-threaded; this keeps resource use predictable and lets users
parallelise over candidate clusterings, which is where parallelism pays off
in a parameter search. Inputs are validated, with informative errors for
empty data, mismatched lengths of \texttt{X} and \texttt{labels}, and \texttt{min\_points\ \textless{}\ 2}.

\textbf{Documentation and tests.} Every exported function has a help page with an
example, a vignette (\texttt{vignette("discoCVI")}) walks through the workflow of
the next section, and the test suite checks the values of all exported
functions (see the implementation section). Bugs and feature requests can be
reported on GitHub (\url{https://github.com/aminentezari/discoCVI/issues}).

\section{A typical workflow}\label{a-typical-workflow}

We illustrate the workflow with two interleaving half-moons and 40 points of
uniform noise.

\begin{verbatim}
library(discoCVI)
library(dbscan)

set.seed(42)
moons <- function(n, sd = 0.08) {
  t1 <- runif(n / 2, 0, pi); t2 <- runif(n / 2, 0, pi)
  rbind(cbind(cos(t1), sin(t1)), cbind(1 - cos(t2), 0.5 - sin(t2))) +
    matrix(rnorm(2 * n, sd = sd), ncol = 2)
}
X <- rbind(moons(600), cbind(runif(40, -1.5, 2.5), runif(40, -1, 1.5)))
\end{verbatim}

\textbf{Scoring a clustering.} We cluster the data with HDBSCAN. Since
\CRANpkg{dbscan} labels noise as \texttt{0} and DISCO expects \texttt{-1}, the labels are
converted first; passing them unchanged would silently treat all noise
points as one extra cluster.

\begin{verbatim}
labels <- hdbscan(X, minPts = 10)$cluster
labels[labels == 0] <- -1L
table(labels)
\end{verbatim}

\begin{verbatim}
#> labels
#>  -1   1   2 
#>  32 304 304
\end{verbatim}

\begin{verbatim}
disco_score(X, labels)
\end{verbatim}

\begin{verbatim}
#> [1] 0.4438417
\end{verbatim}

\begin{verbatim}
disco_score(X, kmeans(X, centers = 2, nstart = 10)$cluster)
\end{verbatim}

\begin{verbatim}
#> [1] 0.1017555
\end{verbatim}

DISCO rates the HDBSCAN result, which follows the moons and separates the
scattered points as noise, higher than the \(k\)-means partition, which cuts
across the moons.

DISCO is considerably stricter than the silhouette coefficient: values
close to 1 are rare, and good clusterings typically score well below what
the silhouette would suggest. Its rules of thumb therefore do not apply,
and DISCO is best used, as here, to compare clusterings of the same data.

\textbf{Choosing a parameter.} Because DISCO needs no ground truth, it can be used
to select the parameter of the clustering algorithm.

\noindent\begin{minipage}{\linewidth}
\begin{verbatim}
score_minpts <- function(m) {
  lab <- hdbscan(X, minPts = m)$cluster
  lab[lab == 0] <- -1L
  disco_score(X, lab)
}
grid <- c(3, 5, 10, 20, 50)
round(sapply(grid, score_minpts), 3)
\end{verbatim}

\begin{verbatim}
#> [1] -0.213  0.445  0.444  0.442  0.365
\end{verbatim}
\end{minipage}

The \texttt{min\_points} argument of \texttt{disco\_score()} is a different parameter: it
is the neighbourhood size DISCO uses to measure density (default 5, as in
the original paper) and need not equal the \texttt{minPts} of the clustering.
Here \texttt{minPts} values from 5 to 20 give almost the same score, so the
choice of 10 above is not critical, whereas \texttt{minPts = 3} is clearly worse.

\begin{figure}[!htbp]
\centering
\includegraphics[width=0.7\linewidth,alt={Scatter plot of two half-moon clusters and scattered noise points, coloured by pointwise DISCO score on a red-yellow-blue scale; most points are light blue and points near the cluster boundaries are paler, close to yellow.}]{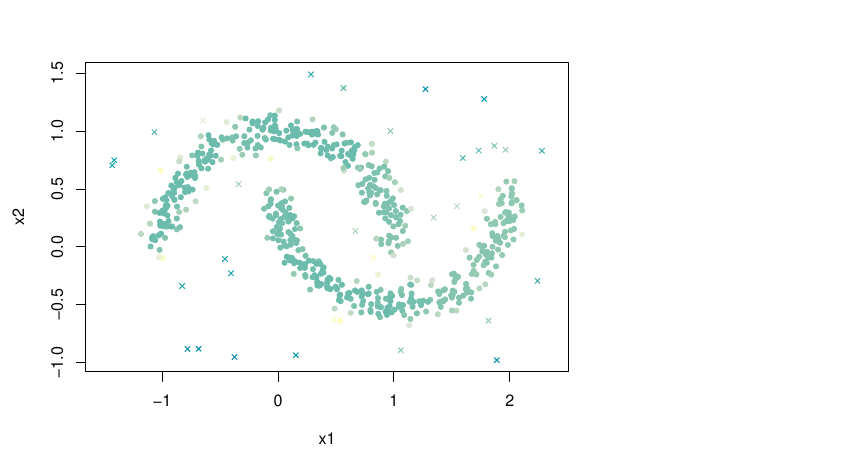} \caption{Pointwise DISCO scores of the HDBSCAN clustering. Colours run from red (score \(-1\)) through yellow (\(0\)) to blue (\(+1\)), so light-blue points are well placed and yellowish points are borderline; crosses mark points labelled as noise.}\label{fig:wf-plot}
\end{figure}

\textbf{Inspecting points.} Pointwise scores show where a clustering is
convincing and where it is not (Figure \ref{fig:wf-plot}). For noise
points, \texttt{p\_noise()} shows which noise labels are least convincing, that is,
which points lie in a dense region or are density-connected to a cluster.

\begin{verbatim}
s  <- disco_samples(X, labels)
nz <- p_noise(X, labels)
noise <- data.frame(point = which(labels == -1),
                    p_sparse = nz$p_sparse, p_far = nz$p_far)
head(noise[order(pmin(noise$p_sparse, noise$p_far)), ], 3)
\end{verbatim}

\begin{verbatim}
#>    point     p_sparse        p_far
#> 1    117 -0.025553586 -0.025553586
#> 31   638 -0.005671184 -0.005671184
#> 2    347  0.025974304  0.025974304
\end{verbatim}

For these points the two components coincide: this happens when the
dc-distance from a noise point to the closest cluster equals its own core
distance, so that both scores compare the same quantities.

\section{Implementation: DISCO without an n-by-n matrix}\label{implementation-disco-without-an-n-by-n-matrix}

\subsection{Key observation}\label{key-observation}

Sort the \(n - 1\) edges of the MST by increasing weight and add them one at a
time, merging the two groups of points they connect (Kruskal order
\citep{kruskal1956}). This produces the single-linkage dendrogram of the
mutual-reachability graph \citep{gower1969mst}: a binary tree whose \(n\) leaves
are the points and whose \(n - 1\) internal nodes are the merges, each with
height \(w_v\) equal to the weight of the corresponding edge.

\textbf{Lemma.} For two points \(x \neq y\), \(d_{dc}(x, y) = w_v\), where \(v\) is the
lowest common ancestor of \(x\) and \(y\) in the merge tree.

\emph{Proof.} The MST path between \(x\) and \(y\) is unique, and \(x\) and \(y\) end up
in the same group exactly when the last edge of that path is added. Since
edges are added in increasing order, this is the largest edge on the path,
which is \(d_{dc}(x, y)\), and it is the edge that creates \(v\). \(\square\)

Now fix a point \(x\) and let \(v_1, \dots, v_r\) be its ancestors, from its
parent up to the root, and \(s_k\) the child of \(v_k\) that does \emph{not} contain
\(x\). The sets of leaves below \(s_1, \dots, s_r\) partition
\(X \setminus \{x\}\), and by the Lemma every point below \(s_k\) is at
dc-distance exactly \(w_{v_k}\) from \(x\). As in the definition, the MST and the
merge tree are built over all \(n\) points, noise included; noise points only
affect which leaves are counted. Writing \(c_C(s)\) for the number of
points of cluster \(C\) below node \(s\), we obtain
\[
S_C(x) = \sum_{y \in C} d_{dc}(x, y) = \sum_{k=1}^{r} w_{v_k}\, c_C(s_k),
\qquad
m_C(x) = \min\{ w_{v_k} : c_C(s_k) > 0 \}.
\]
Both quantities satisfy a simple recursion from the root down: for a node
\(u\) with parent \(v\) and sibling \(s\),
\[
S_C(u) = S_C(v) + w_v\, c_C(s), \qquad
m_C(u) = \begin{cases} \min\{m_C(v), w_v\} & \text{if } c_C(s) > 0,\\
m_C(v) & \text{otherwise,}\end{cases}
\]
with \(S_C(\text{root}) = 0\) and \(m_C(\text{root}) = \infty\). The same
recursion with the number of noise points below \(s\) gives the dc-distance to
the closest noise point, needed in the one-cluster case. Since
\(d_{dc}(x, x) = 0\), \(a(x) = S_C(x) / (|C| - 1)\) and \(b(x)\) follows from
\(S_{C'}(x) / |C'|\); the noise scores use \(m_C(x)\) and the core distances.
Every term of DISCO is therefore available without forming a single
dc-distance explicitly.

\subsection{Algorithm and complexity}\label{algorithm-and-complexity}

The complete procedure is summarised below; internal nodes of the merge tree
are numbered \(n + 1, \dots, 2n - 1\) in the order in which they are created,
so that every parent has a larger number than its children.

\begin{verbatim}
Input : X (n x p), labels (clusters 1..K, noise -1), mu >= 2
Output: pointwise DISCO scores rho(x_1), ..., rho(x_n)

 1  handle edge cases (i) all noise and (ii) one cluster without noise
 2  kappa  <- core distances (distance to the (mu-1)-th other point)
 3  E      <- Prim MST of d_m, computing d_m(x_i, x_j) when needed
 4  sort E by weight; build the merge tree with union-find:
        each edge (i, j, w) creates node v with children
        root(i), root(j) and height w_v = w
 5  bottom-up (v = n+1, ..., 2n-1):
        c_C(v) <- c_C(left) + c_C(right)   for every cluster C
        z(v)   <- z(left)   + z(right)       (number of noise points)
 6  top-down (v = 2n-1, ..., n+1), for both children u of v, sibling s:
        S_C(u) <- S_C(v) + w_v * c_C(s)
        m_C(u) <- min(m_C(v), w_v)  if c_C(s) > 0, else m_C(v)
        m_N(u) <- min(m_N(v), w_v)  if z(s)   > 0, else m_N(v)
 7  cluster points: a = S_own / (|own| - 1),
        b = min_C S_C / |C|   (or m_N in the one-cluster case (iv)),
        rho = (b - a) / max(a, b);  singleton clusters -> 0      (iii)
 8  noise points: rho = min(rho_sparse(kappa), rho_far(m_C, kappa))
 9  set fractions with a zero denominator to 0                    (v)
\end{verbatim}

Step 2 uses \CRANpkg{FNN}. Step 3 is Prim's algorithm \citep{prim1957} written in
C++ via \CRANpkg{Rcpp}; it evaluates \(d_m(x_i, x_j)\) on the fly and needs
\(O(n^2 p)\) time and \(O(n)\) memory. Step 4 uses union by size
\citep{tarjan1975unionfind} and takes \(O(n \log n)\) time. Steps 5 and 6 are
written in R with vectorised operations on \(K \times (2n - 1)\) matrices and
take \(O(nK)\) time and memory. \citet{beer2026disco} state an overall complexity of
\(O(n^2)\) for DISCO, which this algorithm does not change: the running time is
still dominated by the spanning tree. What changes is the memory, which drops
from \(O(n^2)\) to \(O(n(p + K))\), and the constant factor, since no
\(n \times n\) matrix has to be allocated, filled or traversed. The result is
exact: the only differences to a matrix-based computation come from adding
the same distances in a different order.

\subsection{Changes in version 0.2.0 and testing}\label{changes-in-version-0.2.0-and-testing}

\texttt{disco\_score()} and \texttt{disco\_samples()} use the merge-tree algorithm.
\texttt{compute\_dc\_distances()}, which by definition returns an \(n \times n\)
matrix, no longer builds the mutual-reachability matrix and fills the
dc-distance matrix block by block in Kruskal order instead of running a
breadth-first search from every point, and \texttt{p\_cluster()} is vectorised as a
single matrix product. The interface is unchanged.

\textbf{Tests.} The test suite contains the functions of version 0.1.1 as a
reference, with a single change: points in singleton clusters score \(0\), as
explained below, so that both versions follow the published definition. It
compares the values returned by all exported functions over
random configurations: one to eight dimensions, \(\mu\) from 2 to 10, with and
without noise, singleton clusters, one cluster plus noise, mostly-noise
labellings and random labels. All agree to within \(10^{-10}\); observed
differences are of the order of \(10^{-15}\).

\textbf{Singleton clusters.} The DISCO definition assigns score \(0\) to points in
singleton clusters \citep[Section 3.1]{beer2026disco}, as does the silhouette
function of scikit-learn used by the reference implementation. Version 0.1.1
set \(a(x) = 0\) for such points and therefore returned \(1\). Version 0.2.0
follows the definition; labellings without singleton clusters are unaffected.

\section{Performance}\label{performance}

\subsection{Comparison with the previous version}\label{comparison-with-the-previous-version}

\textbf{Setup.} We compared \CRANpkg{discoCVI} 0.1.1 (from CRAN) and 0.2.0 on
synthetic data: three Gaussian clusters in two dimensions plus 5\% uniform
noise, labelled with the generating labels so that only DISCO is measured,
with \(n\) from 500 to 100,000 and \(\mu = 5\). Version 0.1.1 was run up to
\(n = 4{,}000\), beyond which it does not fit comfortably in memory. Every
measurement was a separate \texttt{Rscript} process. Peak memory is the maximum
resident set size of the process (reported by \texttt{/usr/bin/time\ -l}), which
includes R itself (about 67 MB) and the C++ code. Energy was measured with
CodeCarbon \citep{codecarbon} around the whole process, and the median energy of
a baseline process that generates the data without computing DISCO was
subtracted. Each configuration was run five times in random order, after a
warm-up; we report medians. Both versions returned identical scores for
every \(n\). The machine was a 2023 MacBook Pro with an Apple M2 Pro chip and
16 GB of memory, running macOS 27, R 4.5.3, \CRANpkg{FNN} 1.1.4.1,
\CRANpkg{Rcpp} 1.1.2 and CodeCarbon 3.2.8 (Python 3.10), using a single
thread.

\begin{table}[!h]
\centering
\caption{\label{tab:tab-bench}Median DISCO computation time, peak memory of the R process, and energy above the baseline, over five runs. Version 0.1.1 was not run beyond 4,000 points.}
\centering
\resizebox{\ifdim\width>\linewidth\linewidth\else\width\fi}{!}{
\begin{tabular}[t]{rrrrrrrr}
\toprule
n & time 0.1.1 (s) & time 0.2.0 (s) & speed-up & memory 0.1.1 (MB) & memory 0.2.0 (MB) & energy 0.1.1 (mWh) & energy 0.2.0 (mWh)\\
\midrule
500 & 0.249 & 0.011 & 23x & 118 & 91 & 0.26 & 0.04\\
1,000 & 1.004 & 0.022 & 46x & 200 & 92 & 1.32 & 0.16\\
2,000 & 4.309 & 0.053 & 81x & 377 & 89 & 3.99 & 0.13\\
4,000 & 18.554 & 0.134 & 138x & 1222 & 93 & 44.89 & 0.25\\
8,000 & -- & 0.407 &  & -- & 98 & -- & 0.58\\
16,000 & -- & 1.433 &  & -- & 109 & -- & 1.58\\
32,000 & -- & 5.344 &  & -- & 125 & -- & 5.90\\
64,000 & -- & 20.319 &  & -- & 162 & -- & 44.47\\
100,000 & -- & 49.944 &  & -- & 198 & -- & 105.99\\
\bottomrule
\end{tabular}}
\end{table}

\begin{figure}
\includegraphics[width=0.85\linewidth,alt={Log-log plot of computation time against number of points for discoCVI 0.1.1 and 0.2.0; the 0.2.0 line lies about two orders of magnitude lower.}]{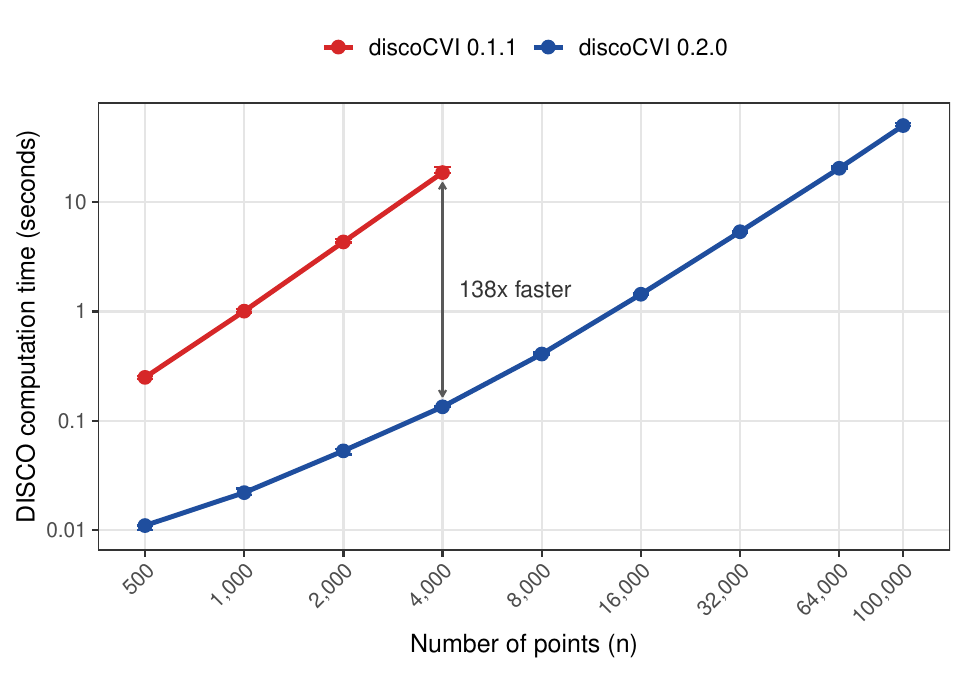} \caption{Time to compute the DISCO score as a function of the number of points (log-log scale). Points are medians and bars the range over five runs.}\label{fig:fig-time}
\end{figure}

\textbf{Time.} Version 0.2.0 is 23 times faster at \(n = 500\) and 138 times faster
at \(n = 4{,}000\) (Table \ref{tab:tab-bench}, Figure \ref{fig:fig-time}).
Both versions scale roughly quadratically; fitted log-log slopes are 2.1 for
version 0.1.1 (500 to 4,000 points) and 1.9 for version 0.2.0 (8,000 to
100,000 points). The gain is a much smaller constant, not a lower complexity
class: 100,000 points take about 50 seconds.

\begin{figure}
\includegraphics[width=0.85\linewidth,alt={Log-log plot of peak memory against number of points; version 0.1.1 rises steeply above one gigabyte, version 0.2.0 stays near 100 megabytes.}]{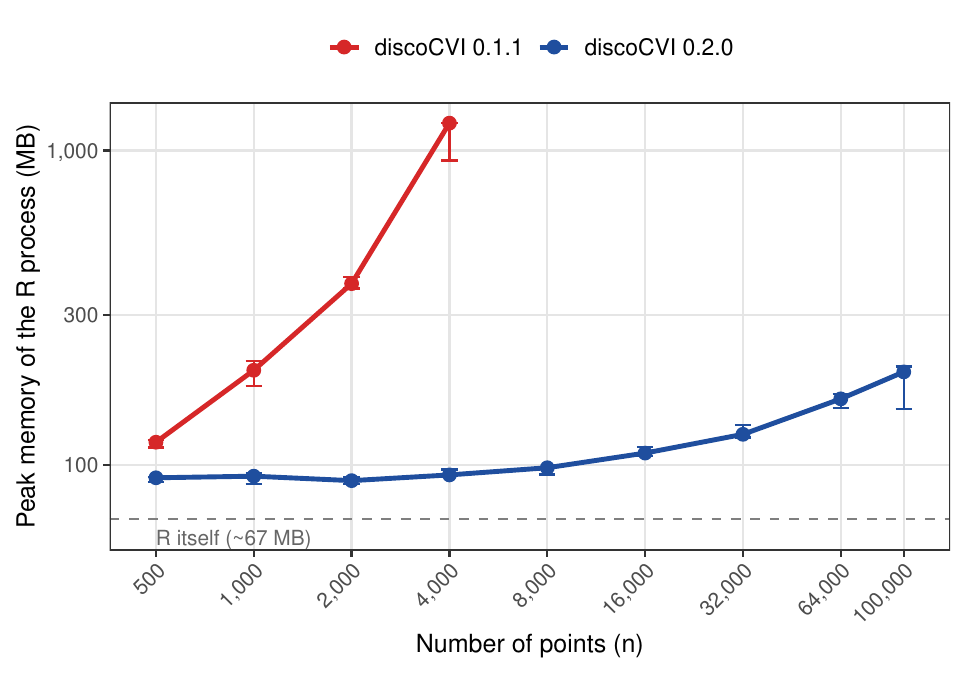} \caption{Peak memory (maximum resident set size) of the whole R process; the dashed line marks the memory used by R itself. Points are medians and bars the range over five runs.}\label{fig:fig-memory}
\end{figure}

\textbf{Memory.} The memory of version 0.1.1 grows rapidly, from 118 MB at
\(n = 500\) to 1,222 MB at \(n = 4{,}000\). Version 0.2.0 stays within about
30 MB of R's own footprint up to \(n = 8{,}000\) and needs 198 MB in total at
\(n = 100{,}000\) (Figure \ref{fig:fig-memory}).

\begin{figure}
\includegraphics[width=0.85\linewidth,alt={Log-log plot of energy against number of points; version 0.1.1 reaches about 45 mWh at 4,000 points, which version 0.2.0 only reaches at 64,000 points.}]{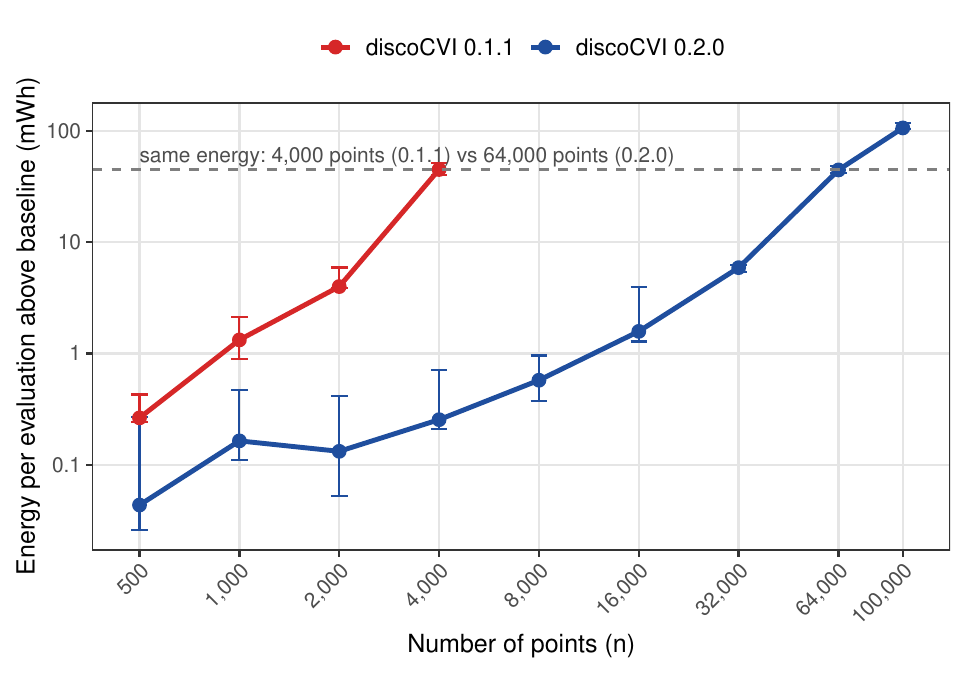} \caption{Energy per DISCO evaluation measured with CodeCarbon, minus the baseline process (log-log scale). The dashed line marks the energy of version 0.1.1 at 4,000 points, which version 0.2.0 reaches only at 64,000 points.}\label{fig:fig-energy}
\end{figure}

\textbf{Energy.} At \(n = 4{,}000\), one evaluation used 44.9 mWh with version
0.1.1 and 0.26 mWh with version 0.2.0 (Figure \ref{fig:fig-energy}). For
long runs the average power was similar for both versions (about 8 W), so
the energy saving follows the time saving. For the same energy budget of
about 45 mWh, version 0.1.1 scores 4,000 points and version 0.2.0 scores
64,000. Energy values for runs shorter than a fraction of a second are close
to the resolution of software-based measurement and should be read as
indicative only.

\subsection{Comparison with the reference implementation}\label{comparison-with-the-reference-implementation}

We also ran the original Python implementation of \citet{beer2026disco} and
\CRANpkg{discoCVI} 0.2.0 on the same data files (the same generator as
above, with \(n\) from 500 to 8,000, plus one dataset of 500 points that
contains a singleton cluster), each in its own process, three times per
dataset. Only the DISCO computation was timed; peak memory is again the
maximum resident set size of the whole process. The reference code uses
three functions from scikit-learn, whose compiled dependencies could not be
loaded on macOS 27; we replaced them by NumPy implementations of the same
formulas, which agreed with scikit-learn to within \(3 \times 10^{-15}\) on test
data on a system where scikit-learn runs. All code specific to DISCO was the
authors' original.

\begin{table}[!h]
\centering
\caption{\label{tab:tab-python}Original Python implementation versus discoCVI 0.2.0 on identical data: median time of the DISCO computation, peak memory of the whole process, and the largest absolute difference between the pointwise scores.}
\centering
\resizebox{\ifdim\width>\linewidth\linewidth\else\width\fi}{!}{
\begin{tabular}[t]{rrrrrrr}
\toprule
n & time Python (s) & time discoCVI (s) & speed-up & memory Python (MB) & memory discoCVI (MB) & max. difference\\
\midrule
500 & 0.032 & 0.011 & 2.9x & 68 & 88 & 1.4e-14\\
1,000 & 0.084 & 0.023 & 3.6x & 103 & 94 & 3.4e-14\\
2,000 & 0.262 & 0.056 & 4.7x & 211 & 93 & 6.4e-14\\
4,000 & 0.919 & 0.130 & 7.1x & 420 & 93 & 1.5e-13\\
8,000 & 3.533 & 0.405 & 8.7x & 1134 & 96 & 2.5e-13\\
\bottomrule
\end{tabular}}
\end{table}

The pointwise scores of the two implementations agree to within
\(3 \times 10^{-13}\) on every dataset (Table \ref{tab:tab-python}), including
the dataset with a singleton cluster (\(7 \times 10^{-14}\)), which confirms
the correction described above. \CRANpkg{discoCVI} is 2.9 times faster at
\(n = 500\) and 8.7 times faster at \(n = 8{,}000\). The memory of the Python
implementation grows with \(n^2\) and reaches 1.1 GB at \(n = 8{,}000\), whereas
\CRANpkg{discoCVI} needs 96 MB, most of which is R itself.

\subsection{Limitations}\label{limitations}

The running time is quadratic in \(n\), so very large datasets (several
hundred thousand points) remain slow even though memory is no longer a
constraint. The package uses a single thread. DISCO is defined for any
dissimilarity \citep{beer2026disco}, whereas \CRANpkg{discoCVI} currently supports
Euclidean distances on numeric data; since the merge-tree passes need only
the spanning tree, other dissimilarities would only require a different
spanning-tree step. The benchmarks were run on one machine with synthetic
two-dimensional data, and energy was measured in software, so absolute
numbers depend on hardware; the comparisons are nevertheless paired (same
data, same machine, interleaved runs), and the equality of the scores is
verified by the test suite rather than by the benchmarks alone.

\section{Discussion}\label{discussion}

The merge-tree formulation removes the memory barrier of DISCO and makes the
index practical for datasets of tens of thousands of points on a laptop,
with results that match both the previous version and the original
implementation. In parameter searches the savings multiply: scoring 20
candidate clusterings of 4,000 points takes about six minutes and 0.9 Wh
with version 0.1.1, and about three seconds and 5 mWh with version 0.2.0.

The remaining cost is the quadratic minimum spanning tree. In low
dimensions, dual-tree Borůvka algorithms compute the Euclidean MST in close
to \(O(n \log n)\) time \citep{march2010boruvka} and are used to accelerate HDBSCAN
\citep{mcinnes2017accelerated}; the same idea applies to mutual-reachability
distances and would combine directly with the merge-tree passes described
here. When DISCO is computed for an HDBSCAN result with the same value of
\(\mu\), the spanning tree built by the clustering itself could also be
reused. Both are planned for future versions.

\section{Summary}\label{summary}

\CRANpkg{discoCVI} provides the DISCO index in R through a small interface
of plain functions that works with labels from any clustering algorithm.
Since version 0.2.0 it computes DISCO exactly from the merge tree of the
mutual-reachability minimum spanning tree, without storing any \(n \times n\)
matrix. It agrees with the original Python implementation, follows the
published definition for singleton clusters, is up to two orders of
magnitude faster than its previous version and up to nine times faster than
the reference implementation, and scores 100,000 points in under a minute
with less than 200 MB of memory. The package is available from CRAN; the
benchmark scripts and data are in its source repository at
\url{https://github.com/aminentezari/discoCVI}.

\section*{Acknowledgements}\label{acknowledgements}

The author thanks Anna Beer, Lena Krieger and Pascal Weber for the original
Python implementation of DISCO, and Davide Chicco for supervising the
development of version 0.1 of \CRANpkg{discoCVI}.

\bibliographystyle{plainnat}
\bibliography{disco-references}

\end{document}